\documentclass[letterpaper, 10 pt, conference]{ieeeconf}  

\IEEEoverridecommandlockouts                              

\usepackage{graphics} 
\usepackage{epsfig} 
\usepackage{times} 
\usepackage{amsmath} 
\usepackage{amssymb}  
\usepackage{tabularx, booktabs}
\usepackage{subcaption}
\usepackage{lipsum}
\usepackage{mathtools}
\usepackage{gensymb}

\newcolumntype{Y}{>{\centering\arraybackslash}X}

\newcommand{\Exp}{\text{Exp}}
\newcommand{\R}{\mathbb{R}}
\newcommand{\argdot}{\makebox[1ex]{\textbf{$\cdot$}}}
\newcommand{\ehat}[1]{#1^\wedge}
\newcommand{\Var}{\text{Var}}

\newcommand{\Stwo}{{\mathbb{S}^2}}

\title{\LARGE \bf Leg Exoskeleton Odometry using a Limited FOV Depth Sensor}

\author{Fabio Elnecave Xavier$^{1,2}$, Matis Viozelange$^1$, Guillaume Burger$^1$, Marine Pétriaux$^1$,\\Jean-Emmanuel Deschaud$^2$ and François Goulette$^{2,3}$
\thanks{$^1$Wandercraft, 75004 Paris, France}%
\thanks{$^2$Centre for Robotics, Mines Paris, PSL University, 75006 Paris, France}%
\thanks{$^3$U2IS, ENSTA Paris, Institut Polytechnique de Paris, 91120 Palaiseau, France}%
\thanks{\tt\small fabio.elnecave\_xavier@minesparis.psl.eu}%
}

\begin{document}

\maketitle
\thispagestyle{empty}
\pagestyle{empty}

\begin{abstract}

For leg exoskeletons to operate effectively in real-world environments, they must be able to perceive and understand the terrain around them. However, unlike other legged robots, exoskeletons face specific constraints on where depth sensors can be mounted due to the presence of a human user. These constraints lead to a limited Field Of View (FOV) and greater sensor motion, making odometry particularly challenging. To address this, we propose a novel odometry algorithm that integrates proprioceptive data from the exoskeleton with point clouds from a depth camera to produce accurate elevation maps despite these limitations. Our method builds on an extended Kalman filter (EKF) to fuse kinematic and inertial measurements, while incorporating a tailored iterative closest point (ICP) algorithm to register new point clouds with the elevation map. Experimental validation with a leg exoskeleton demonstrates that our approach reduces drift and enhances the quality of elevation maps compared to a purely proprioceptive baseline, while also outperforming a more traditional point cloud map-based variant.

\end{abstract}


\section{INTRODUCTION}

Self-balancing leg exoskeletons enable individuals with motor disabilities to walk completely hands-free. Currently, their use is restricted to hospitals and rehabilitation centers, where they are operated under strict supervision. To extend their application beyond these clinical settings and make them suitable for personal mobility, these devices must be able to navigate safely in less controlled environments. This requires the development of a perception system that allows the exoskeleton to understand the terrain and the obstacles in its surroundings, so that it can plan its movements accordingly.

Most previous works on perception for exoskeletons focused on tasks such as classifying the terrain type \cite{al2022depth} or detecting staircases \cite{kurbis2022stair}. However, a more general approach that has been successfully adopted by different legged robots is to construct local elevation maps that represent the geometry of the terrain around them \cite{mastalli2020motion}. This, in turn, requires the robot to be equipped with a depth sensor and to run an odometry algorithm to estimate its trajectory as it walks \cite{fankhauser2018probabilistic}.

\begin{figure}[t]
    \centering
    \includegraphics[width=0.95\linewidth]{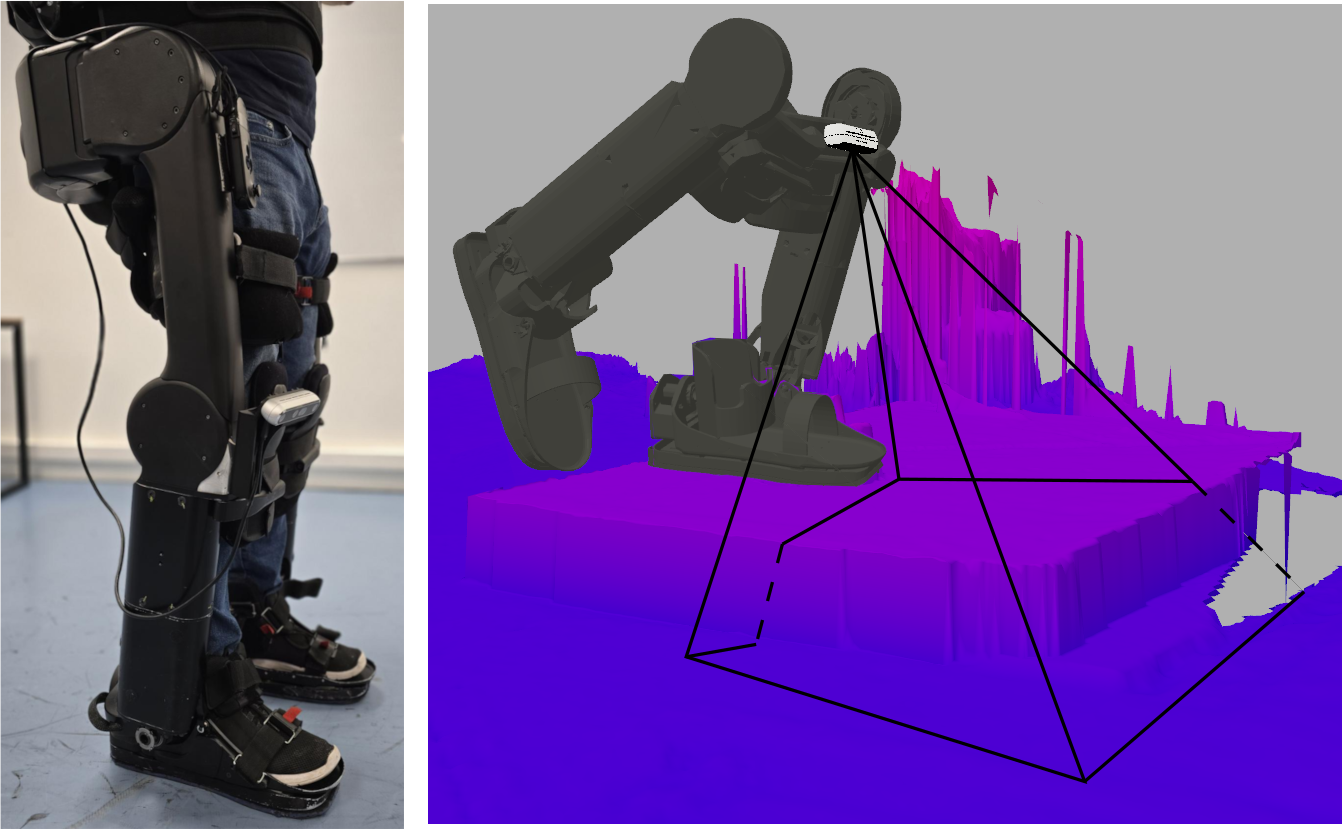}
    \caption{On the left, photo of the personal exoskeleton from Wandercraft with a depth camera mounted above its right knee. On the right, 3D view showing the limited FOV of the camera at this position as the exoskeleton climbs a step.}
    \vspace{-3ex}
    \label{fig:exo_fov}
\end{figure}

An important difference between leg exoskeletons and other legged robots lies in the constraints on where a depth sensor can be mounted. Since there is always a person wearing the exoskeleton, this sensor must be placed sufficiently below the waist level (e.g. slightly above the knee, as illustrated in Fig. \ref{fig:exo_fov}) to avoid the risk of obstruction by their body or arms. At this location, the sensor has a more limited Field Of View (FOV) and is subjected to higher velocities and accelerations compared to a sensor mounted on the head of a humanoid robot or on the floating base of a quadruped.

These conditions can have a significant impact on the performance of odometry algorithms that rely on visual information. The restricted FOV often results in the observed region possessing a very simple geometry, such as a flat floor with few distinguishing features, which can lead to degenerate cases for point cloud registration methods. Limited and uniform visual data may also compromise the detection of feature points for visual odometry, while the rapid motion of the sensor can disrupt their tracking.

Although the experiments were not conducted with an exoskeleton, the impact of a reduced FOV under similar circumstances was evaluated in \cite{wang2024exosense}. The performance of the state-of-the-art visual-inertial odometry system \cite{wisth2022vilens} was tested using a multi-camera setup attached to the leg of a human walking at normal speed. The authors report that, with standard FOV cameras, the trajectory estimates exhibited significant drift, or the feature tracking even failed. This issue was mitigated by using wide FOV cameras, which reduced the drift rates significantly, even though they remained higher than those typically achieved with handheld devices.

In this paper, we present a new odometry algorithm that is able to produce accurate elevation maps under these challenging conditions. Our method combines data from the exoskeleton's proprioceptive sensors with point clouds captured by a depth camera. We build on the proprioceptive state estimator described in \cite{xavier2023multi}, which fuses kinematic and inertial measurements using an extended Kalman filter (EKF). While this estimator is not affected by the placement of the camera, the elevation maps generated from its trajectory estimates suffer from inaccuracies and lack smoothness due to inherent drift in the proprioceptive algorithm and orientation errors in the sensor poses.

To address these issues, we integrate point cloud registration into the state estimation process. Specifically, we adapt the traditional iterative closest point (ICP) algorithm \cite{besl1992method} to align new point clouds with the existing elevation map. This approach not only eliminates the need for a separate point cloud map but also improves computational efficiency. Furthermore, to handle degenerate cases where point cloud registration might be unreliable, we introduce a novel model for computing the ICP covariance matrix, which accurately encodes directions lacking geometric features in the environment. Our experiments with a leg exoskeleton demonstrate that our approach significantly outperforms a purely proprioceptive odometry baseline, effectively reducing drift and improving the quality of elevation maps. Moreover, our elevation map-based ICP variant proves to be more effective for this application than doing ICP using point cloud maps.

The remainder of this paper is organized as follows. In Section \ref{sec:icp} we present our point-cloud-to-elevation-map ICP variant and our model for computing its covariance matrix. Section \ref{sec:odometry_mapping} details how this algorithm is integrated with the proprioceptive state estimator to form a cohesive odometry and mapping pipeline. Finally, Section \ref{sec:results} discusses the experimental results.


\section{POINT-CLOUD-TO-ELEVATION-MAP ICP}
\label{sec:icp}

Point clouds and elevation maps represent the three-dimensional geometry of the environment in different ways. As such, certain adaptations are required to apply the ICP algorithm between these two types of data. Section \ref{sec:icp_transform} discusses how to modify this algorithm to compute the rigid transformation that best aligns a point cloud with an elevation map. The calculation of a covariance matrix that encodes the uncertainty of this transformation is detailed in Section \ref{sec:icp_covariance}. Our discussion is restricted to the point-to-plane variant of ICP \cite{rusinkiewicz2001efficient}, as surface normal information is necessary for the covariance computation.

\subsection{Computing the rigid transformation}
\label{sec:icp_transform}

We follow the standard sequence of steps in the ICP algorithm. After downsampling the point cloud, we iteratively estimate correspondences, compute normal vectors, and solve the optimization problem until convergence. The details of each operation are described below.

\vspace{1ex}
\subsubsection{Point cloud downsampling}

Elevation maps discretize the horizontal plane into a grid of cells, each containing a single estimate of the terrain's elevation at that specific location. This means that, unlike point clouds, they cannot represent vertical surfaces or structures. Our point cloud downsampling procedure is designed to only keep points that can be represented in the elevation map. Consistent with standard ICP practice, we assume that the point cloud has been pre-transformed using an initial estimate of the camera pose, ensuring it is roughly aligned with the elevation map. We then organize the points according to the cell in the map corresponding to their location, retaining only the highest point within each cell and discarding the others. This approach not only enhances the robustness of data association by eliminating points from vertical structures but also improves convergence speed by significantly reducing the number of points used in the next steps.

\vspace{1ex}
\subsubsection{Correspondence estimation}

Each point $ q $ in the downsampled point cloud is matched with its closest point $ q' $ in the elevation map. This matching process can be performed very efficiently by first locating the cell that lies at the location of $ q $ and then restricting the search to its 3x3 pixel neighborhood. We interpret each cell in this neighborhood as a 3D point, with the $ x $ and $ y $ coordinates of the center of the cell and the $ z $ coordinate equal to the cell's elevation. The closest point $ q' $ is then selected as the one with the smallest Euclidean distance to $ q $ among these candidates. We only keep pairs of points whose distance is inferior to a threshold $ D_{max} $.

\vspace{1ex}
\subsubsection{Normal vector computation}

The normal vector to the surface defined by a point cloud at a given location is typically calculated by fitting a plane to the points within a small neighborhood. However, this method is not well suited for elevation maps due to their lower point density. Instead, we adopt a different approach based on image derivatives.

An elevation map can be viewed as a sampling of a continuous surface defined by the function $ z = f(x, y) $. This surface can be equivalently described by the level set $ F(x, y, z) = 0 $, where $ F(x, y, z) = z - f(x, y) $. The normal vector at a point $ \left(x, y, f(x, y) \right) $ is thus parallel to the gradient
\begin{equation}
    \nabla F = \left(-\frac{\partial f}{\partial x}, -\frac{\partial f}{\partial y}, 1 \right)
\end{equation}
evaluated at that point. Consequently, computing the normal vector reduces to computing the partial derivatives of the function $ f $ which represents the elevations of the map. We estimate these derivatives using the Sobel operator, scaled so that they are taken with respect to the $ x $ and $ y $ coordinates expressed in meters rather than in pixels.

We discard all normals that form an angle with the vertical direction larger than a threshold $ \phi_{max} $. This heuristic is justified by the fact that vertical surfaces cannot be accurately represented in an elevation map. Normals that deviate significantly from the vertical are likely computed from a 3x3 neighborhood of cells that cannot be well approximated by a plane, such as at an elevation discontinuity.

\vspace{1ex}
\subsubsection{Robust optimization}

After the previous steps, we obtain two sets of $ N $ corresponding points $ \{q_k\} $ and $ \{q_k'\} $, along with their associated normal vectors $ \{n_k\} $. The rigid transform $ (R, p) \in \text{SO(3)} \times \R^3 $ that best aligns these points is then computed by minimizing the point-to-plane metric
\begin{equation}
    \label{eq:robust_cost}
    J(R, p) = \sum_{k=1}^{N} \rho \left( {n_k}^T \left[ R q_k + p - q_k' \right] \right)
\end{equation}
where $ \rho(\argdot) $ represents the Cauchy robust cost function \cite{babin2019analysis}.

Using the linear approximation $ R = \Exp{(\theta)}  \approx I + \ehat{(\theta)} $\footnote{We define $ \ehat{(a)} $ as the skew-symmetric matrix such that $ \ehat{(a)} b = a \times b $ for all $ a,b \in \R^3 $. $ \Exp{(\theta)} = \exp{(\ehat{\theta})} $ is the rotation matrix associated to the rotation vector $ \theta $.} and an iteratively reweighted least squares approach \cite{bergstrom2014robust}, we transform \eqref{eq:robust_cost} into a sequence of linear least squares problems of the form
\begin{equation}
    J(\tau) = \sum_{k=1}^{N} \left( {a_k}^T \tau - b_k \right)^2 = ||A \tau - b ||^2
\end{equation}
where
\begin{equation}
    \label{eq:definitions_least_squares}
    \tau = 
    \begin{pmatrix}
        \theta \\ p
    \end{pmatrix}
    ,~~
    a_k = \sqrt{w_k}
    \begin{pmatrix}
        q_k \times n_k \\ n_k
    \end{pmatrix}
    ,~~
    b_k = \sqrt{w_k} {n_k}^T(q_k' - q_k)
\end{equation}
and $ w_k $ are the weights that approximate this robust function \cite{babin2019analysis}.
The optimal value $ \tau^* $ is given by the standard least-squares solution
\begin{equation}
    \label{eq:least_squares_solution}
    \tau^* = (A^T A)^{-1} A^T b
\end{equation}

\subsection{Estimating the covariance matrix}
\label{sec:icp_covariance}

Several methods have been proposed to compute the covariance of the transformation estimated by the ICP algorithm. While Monte Carlo simulations \cite{buch2017prediction} and data-driven techniques \cite{landry2019cello} can yield very accurate results, applications requiring real-time covariance estimation typically rely on closed-form expressions based on the shape of the ICP cost function \cite{censi2007accurate}. Despite its simplicity, the analysis in \cite{bonnabel2016covariance} shows that this latter approach can effectively capture the underconstrained directions in the environment, provided that the point-to-plane variant of ICP is used.

To derive a closed-form expression for this covariance matrix, we consider the linear least squares solution \eqref{eq:least_squares_solution}. The traditional assumption in linear regression is that the matrix $ A $ is known exactly, while the elements in $ b $ are corrupted by independent, identically distributed noise terms with variance $ \sigma_b^2 $. This leads to the well-known formula $ \Var(\tau^*) = \sigma_b^2 (A^T A)^{-1} $ \cite{bonnabel2016covariance}. However, assuming $ A $ to be known disregards the fact that the estimated normal vectors $ n_k $ are also noisy, which can result in overconfident variance estimations in the underconstrained directions.

To address this, we propose a different model that accounts for the uncertainty in the normal vectors. We start by considering how the solution \eqref{eq:least_squares_solution} changes when both $ A $ and $ b $ are perturbed by additive noise terms $ \delta A $ and $ \delta b $. The perturbed solution $ \tau^* + \delta \tau $ is
\begin{equation}
    \tau^* + \delta \tau = [(A + \delta A)^T (A + \delta A)]^{-1} (A + \delta A)^T (b + \delta b)
\end{equation}
Linearizing this equation with respect to $ \delta A $ and $ \delta b $, we obtain the approximation\footnote{This assumes that $ (A^T A)^{-1} A^T b \approx 0 $, which is natural since this calculation is performed after the ICP has converged.}
\begin{equation}    
    \begin{split}
        \delta \tau &\approx (A^T A)^{-1}(A^T \delta b + \delta A^T b) \\
        &= (A^T A)^{-1} \left(A^T \delta b + \sum_{k=1}^{N} b_k \delta a_k \right)
    \end{split}
\end{equation}
Under the assumption that the perturbations $ \delta a_k $ and $ \delta b $ are all independent of each other, and keeping the previous model for the noise of $ b $, the expression for the covariance matrix of $ \tau^* $ becomes
\begin{equation}
    \Var(\tau^*) = \sigma_b ^2 (A^T A)^{-1} + (A^T A)^{-1} \left[\sum_{k=1}^{N} b_k^2 \Var(a_k) \right] (A^T A)^{-1}
\end{equation}

To model the noise in the matrix $ A $, we evaluate how perturbations in the normal vectors $ n_k $ translate to perturbations in the vectors $ a_k $. Since the normal vectors lie on $ \Stwo $, we express their perturbations using the $ \oplus $ operator for this manifold, as defined in \cite{xavier2023multi}. From \eqref{eq:definitions_least_squares}, we obtain
\begin{equation}
    \label{eq:perturbation_a}
    a_k + \delta a_k = \sqrt{w_k}
    \begin{pmatrix}
        q_k \times (n_k \oplus \delta n_k) \\ n_k \oplus \delta n_k
    \end{pmatrix}
\end{equation}
We consider the $ \delta n_k $ terms to be independent, with variances $ \Var(\delta n_k) = \sigma_n^2 I $. Then, following the same procedure as in the previous paragraph, the covariance matrices $ \Var(a_k) $ are computed from the linear approximation of \eqref{eq:perturbation_a}. We omit the detailed calculations and present only the final result
\begin{equation}
    \Var(a_k) = \sigma_n^2 w_k
    \begin{pmatrix}
        \ehat{(q_k)} \\ I
    \end{pmatrix}
    ~
    (I-n_k n_k^T)
    \begin{pmatrix}
        -\ehat{(q_k)} & I
    \end{pmatrix}
\end{equation}


\section{ODOMETRY AND MAPPING}
\label{sec:odometry_mapping}

We describe in this section our odometry and mapping pipeline, combining the proprioceptive state estimator from \cite{xavier2023multi} with point cloud registration. The proprioceptive estimator runs an EKF at a fixed frequency, independent of the camera's frame rate. Among other variables, it estimates the pose of multiple IMUs within the exoskeleton's structure. We assume that the camera pose relative to one of these IMUs -- the \textit{reference IMU} -- is either constant or can be accurately determined from the kinematic model at all times. Both the EKF and the elevation map share a fixed reference frame, referred to as the \textit{world frame}.

Upon receiving a new point cloud, we register it to the elevation map, producing an updated estimate of the camera pose. An additional EKF correction step is then performed, using as measurement the result of the ICP. This operation refines the estimated pose of the reference IMU, yielding the \textit{a posteriori} estimate of the camera pose, which is then used to update the elevation map.

\subsection{Updating the state estimator}

Following the framework and the notation defined in \cite{xavier2023multi}, all that is needed to update the EKF is to mathematically define the measurement $ y_{ICP} $ of the camera pose obtained via ICP as a function of the state of the filter. For such, let  $ (R_I, p_I) \in \text{SO(3)} \times \R^3 $ be the pose of the reference IMU in the world frame. The EKF represents this pose as
\begin{equation}
    \label{eq:imu_pose}
    (R_I, p_I) = (\hat{R}_I \oplus \delta \theta_I, \hat{p}_I + \delta p_I)
\end{equation}
where $ (\hat{R}_I, \hat{p}_I) $ is the current estimate and $ (\delta \theta_I, \delta p_I) \in \R^3 \times \R^3 $ are the error state variables. Using the camera pose relative to the reference IMU $ (\prescript{I}{}{R_C}, \prescript{I}{}{p_C}) $, the camera pose in the world frame $ (R_C, p_C) $ can be computed as
\begin{equation}
    \label{eq:camera_pose}
    (R_C, p_C) = (R_I \prescript{I}{}{R_C}, p_I + R_I \prescript{I}{}{p_C})
\end{equation}

We define the measurement $ y_{ICP} $ as
\begin{equation}
    \label{eq:icp_measurement}
    y_{ICP} = (R_C \oplus \nu^R_{ICP},~ p_C + \nu^p_{ICP})
\end{equation}
with $ \nu_{ICP} = [{\nu^R_{ICP}}^T ~ {\nu^p_{ICP}}^T]^T $ representing the uncertainty in the point cloud registration. Its covariance matrix is
\begin{equation}
    \Var(\nu_{ICP}) = M \Var(\tau^*) M^T, ~~ M=
    \begin{pmatrix}
        {\hat{R}_C}^T & 0 \\
        -\ehat{(\hat{p}_C)} & I
    \end{pmatrix}
\end{equation}
where $ (\hat{R}_C, \hat{p}_C) $ is the camera pose according to ICP. Expanding \eqref{eq:icp_measurement} yields
\begin{equation}
    \label{eq:icp_measurement_expanded}
    \begin{split}
        y_{ICP} = & \left( \left[(\hat{R}_I \oplus \delta \theta_I) \prescript{I}{}{R_C}\right] \oplus \nu^R_{ICP}, \right.\\
        & \left. \hat{p}_I + (\hat{R}_I \oplus \delta \theta_I) \prescript{I}{}{p_C} + \delta p_I + \nu^p_{ICP} \right)
    \end{split}
\end{equation}
from which we derive the equation of the linearized measurement
\begin{equation}
    \delta y_{ICP} =
    \begin{pmatrix}
        \prescript{I}{}{R_C}^T \delta \theta_I + \nu^R_{ICP} \\
        -\hat{R}_I \ehat{(\prescript{I}{}{p_C})} \delta \theta_I + \delta p_I + \nu^p_{ICP}
    \end{pmatrix}
\end{equation}

\subsection{Updating the elevation map}

After updating the state of the Kalman filter, we proceed to adding the latest point cloud to the elevation map. Since the elevation of a cell is intended to represent the maximum height at that location, we take as input the downsampled point cloud computed in the first step of the ICP algorithm, transformed into the world frame using the \textit{a posteriori} camera pose. For each point in this point cloud, let $ z $ represent its coordinate in the vertical direction and $ \sigma_z^2 $ the variance representing the uncertainty of this measurement, which we take to be proportional to the square of the distance from the point to the camera. We then identify the corresponding cell in the map, with its current elevation and variance denoted by $ h $ and $ \sigma_h^2 $. The updated elevation $ h^+ $ and variance $ {\sigma_h^2}^+ $ for this cell are computed as follows.

\begin{itemize}
    \item If there is no prior elevation estimate for the cell, the point is simply added to the map:
    \begin{equation}
        h^+ = z, ~~ {\sigma_h^2}^+ = \sigma_z^2
    \end{equation}
    \item If $ z $ lies within the interval $ [h - 2 \sigma_h, h + 2 \sigma_h] $, the current estimate is merged with the new measurement:
    \begin{equation}
        h^+ = \frac{\sigma_h^2 z + \sigma_z^2 h}{\sigma_h^2 + \sigma_z^2}, ~~ {\sigma_h^2}^+ = \frac{\sigma_h^2 \sigma_z^2}{\sigma_h^2 + \sigma_z^2}
    \end{equation}
    \item If $ z $ lies outside this interval, the elevation is not updated, but its variance is increased:
    \begin{equation}
        h^+ = h, ~~ {\sigma_h^2}^+ = \sigma_h^2 + \lambda (z - h)^2
    \end{equation}
\end{itemize}
The rationale behind the last update rule is that measurements falling outside the current confidence interval should not immediately alter the map; instead, the interval size is incrementally increased, requiring several measurements before the elevation is updated. The parameter $ \lambda $ controls the rate of this adjustment.

This is similar to the map update procedure described in \cite{kleiner2007real} and \cite{fankhauser2018probabilistic}, with one important distinction: in those methods, a height measurement that falls outside but above the current confidence interval directly replaces the current elevation estimate. Since these methods do not downsample the point clouds before adding them to the map, this rule ensures that the elevation reflects the highest point in the cell. However, we find that downsampling the point cloud to retain only the highest point per cell, and then uniformly applying the same rule to all points outside the confidence interval, results in more accurate elevation maps.


\section{EXPERIMENTAL VALIDATION}
\label{sec:results}
\begin{figure*}[t!]
    \centering
    \includegraphics[width=0.98\linewidth]{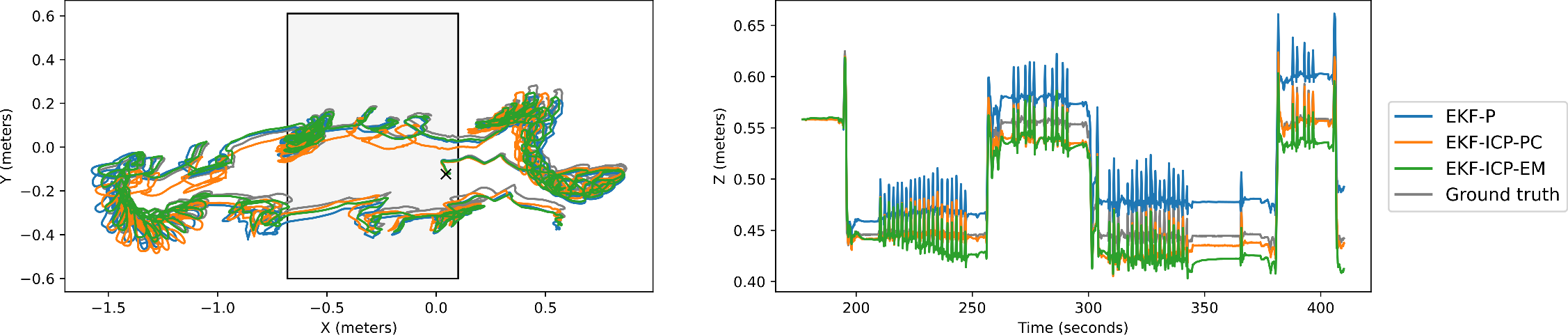}
    \caption{Trajectories from experiment 1. The rectangle in the XY plot indicates the location of the box. The starting point of the trajectory is marked with an ``x''.}
    \label{fig:trajectories_1}
\end{figure*}

\begin{figure*}[t!]
    \centering
    \includegraphics[width=0.97\linewidth]{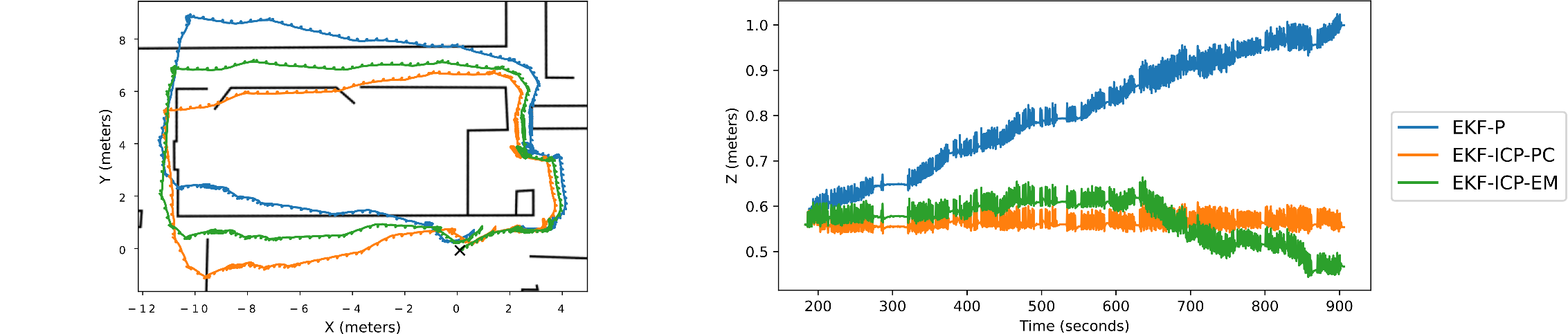}
    \caption{Trajectories from experiment 2. The XY plot is overlaid on a floor plan of the building. The starting point of the trajectory is marked with an ``x''.}
    \vspace{-5mm}
    \label{fig:trajectories_2}
\end{figure*}

\subsection{Methodology}

We tested our method using data collected with the personal exoskeleton developed by Wandercraft, which is instrumented with 12 joint encoders and 7 low-cost MEMS IMUs, and walks at a speed of 0.25 m/s. A RealSense D455 depth camera was mounted above its right knee, rigidly linked to the right shank, as shown in Fig. \ref{fig:exo_fov}. We conducted two experiments representing situations in which the elevation map is needed, namely climbing and descending a step in experiment 1 and walking through doorways in experiment 2. All odometry and mapping results were generated in post-processing, since the algorithm has not yet been implemented for onboard execution on the exoskeleton. The elevation maps measure 4 m by 4 m, with a resolution of 1 cm. We employed the values $ D_{max} $ = 5 cm, $ \phi_{max} $ = 20$\degree$ and $ \lambda = 0.025 $.

Experiment 1 was performed in a square area measuring approximately 4 m per side. A rectangular box, measuring 1.2 m by 0.8 m with a height of 11 cm, was placed in the center and used to represent a step, as depicted in the 3D view of Figure \ref{fig:exo_fov}. Starting from the top of the box, the exoskeleton completed a few straight-line passes in the room, descending from the box three times and climbing it twice. Ground truth was obtained using an OptiTrack motion capture system. For experiment 2, the exoskeleton followed a roughly rectangular path within a building, measuring about 15 m by 6 m, passing through three doorways along the way. Ground truth data is not available for this experiment due to the impossibility of setting up motion capture equipment in this environment.

To validate that our approach successfully reduces drift and produces smoother elevation maps compared to a purely proprioceptive odometry baseline, we compare our results to the state estimator from \cite{xavier2023multi}. Additionally, we also evaluate a variant of our method in which we use a standard point-to-plane ICP algorithm to register point clouds to a point cloud map, but keeping our new model for the covariance matrix. This is intended to determine whether our choice of performing ICP against an elevation map -- a simpler representation of a 3D environment -- significantly impacts the performance of the odometry algorithm. We will refer to these three methods as EKF-P (the purely Proprioceptive EKF), EKF-ICP-EM (our approach, with ICP on Elevation Maps) and EKF-ICP-PC (the Point Cloud map variant).

The point cloud map from EKF-ICP-PC is constructed using a voxel hash map, which accelerates nearest-point searches, and is updated using the procedure described in \cite{vizzo2023kiss}. Normals at each point are computed by fitting a plane to its neighborhood using RANSAC, and only those points with a planarity descriptor \cite{deschaud2018imls} above a threshold of 0.6 are considered valid and retained.

\subsection{Results}

\subsubsection{Experiment 1 -- climbing and descending a step}

The odometry results from experiment 1 are presented in Fig. \ref{fig:trajectories_1}, alongside the ground truth trajectory. The trajectory estimate from EKF-P exhibits an upward drift; meanwhile, both ICP-based methods display a downward drift, but with a smaller overall error. Although EKF-ICP-EM performs slightly worse than EKF-ICP-PC in the vertical direction, its XY trajectory is the closest to the ground truth.

The elevation maps at the end of the experiment are shown in the left column of Fig. \ref{fig:elevation maps}. The map from EKF-P reveals an irregular surface with height variations and inclinations that do not correspond to the actual terrain. In contrast, the maps produced by both ICP-based methods are significantly smoother and more representative of the true environment.

Table \ref{tab:metrics} summarizes the absolute trajectory error (ATE) and relative error (RE) metrics \cite{zhang2018tutorial} for the all methods. We also report under the acronym EKF-ICP-EM* the metrics obtained when running the EKF-ICP-EM algorithm without our new model for the ICP covariance matrix. The results demonstrate that EKF-ICP-EM outperforms all alternatives.

\subsubsection{Experiment 2 -- walking through doorways}

The trajectory estimates for experiment 2 are illustrated in Fig. \ref{fig:trajectories_2}. Although ground truth data is not available, a qualitative assessment of the methods' performance can still be made. First, since the exoskeleton walked on flat ground without encountering any steps or slopes, its vertical position should remain approximately constant. Additionally, the accuracy of the trajectory in the XY plane can be evaluated by overlaying the estimates on the building's floor plan.

Similar to experiment 1, EKF-P exhibits a consistent upward drift, resulting in larger vertical errors compared to both ICP-based estimators. While EKF-ICP-PC maintains a highly accurate Z estimate, EKF-ICP-EM shows a more pronounced downward drift during two intervals, both following 90-degree turns. However, when comparing the XY trajectories with the floor plan, EKF-ICP-EM outperforms the other two methods, being the only one to follow the corridor layout and cross all doorways at the correct locations.

\begin{figure}[h]
    \centering
    
    \begin{tabular}{cc}
        \vspace{1mm}
        \textbf{Experiment 1} & \textbf{Experiment 2} \\
        \includegraphics[width=0.20\textwidth]{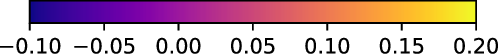} &
        \includegraphics[width=0.20\textwidth]{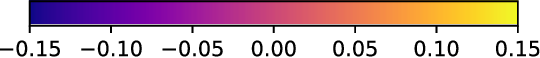} \\\\
        \includegraphics[width=0.20\textwidth]{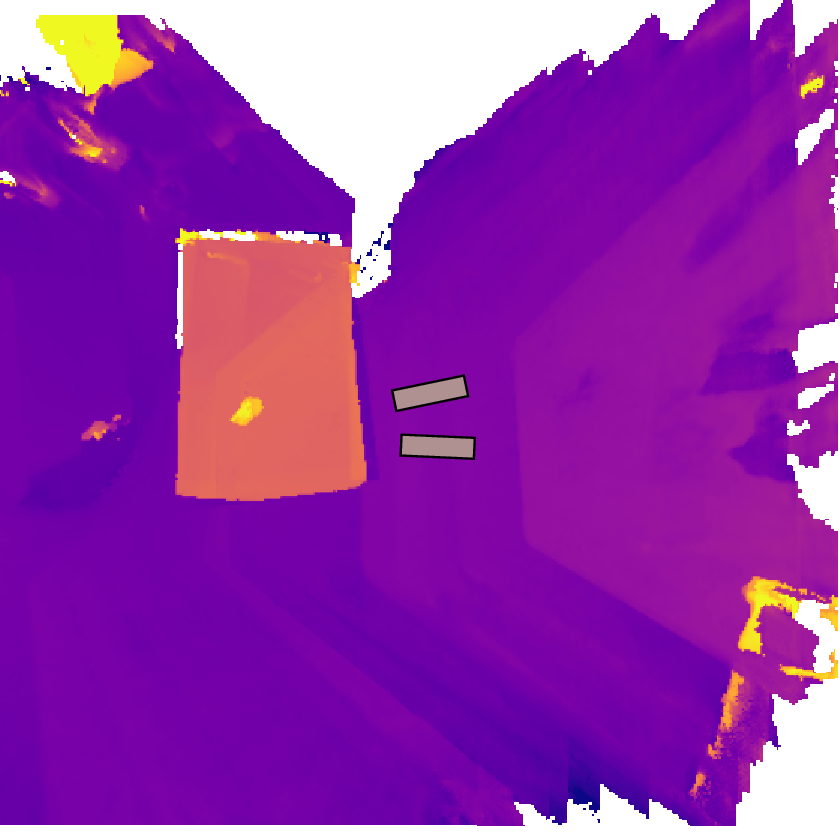} &
        \includegraphics[width=0.20\textwidth]{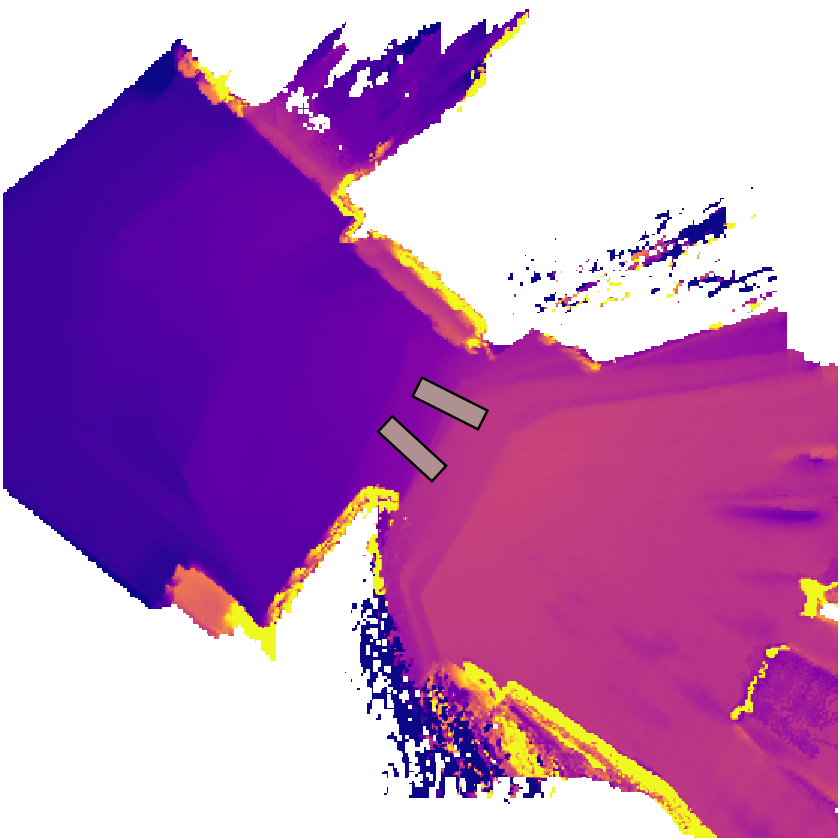}
    \end{tabular}
    \vspace{-2mm}
    \caption*{\small \bf EKF-P}
    \vspace{2mm} 

    \begin{tabular}{cc}
        \includegraphics[width=0.20\textwidth]{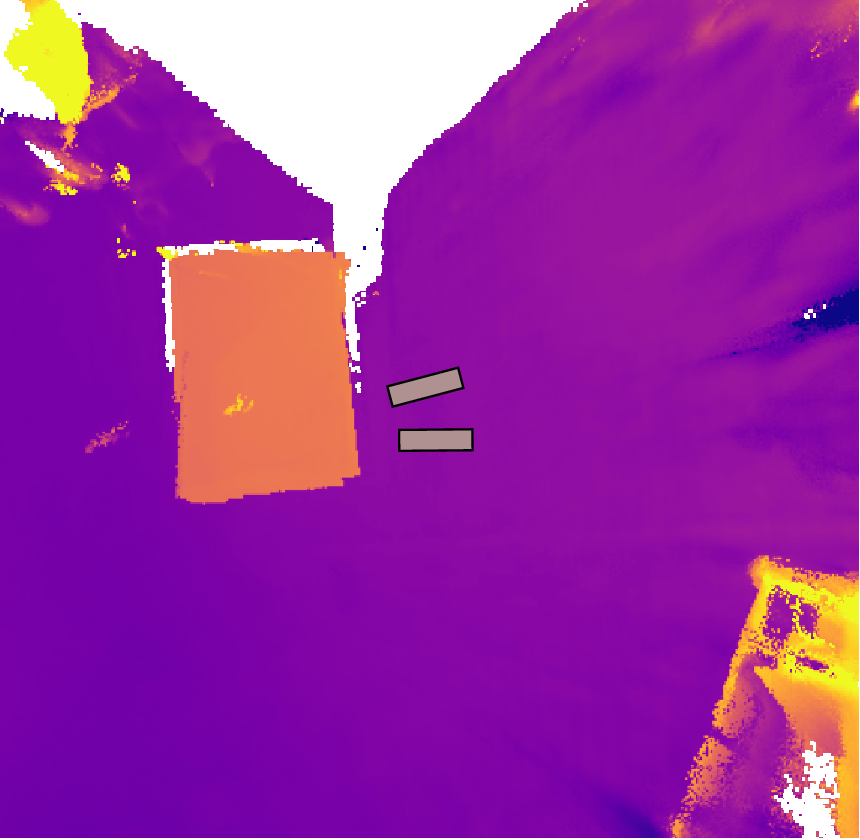} &
        \includegraphics[width=0.20\textwidth]{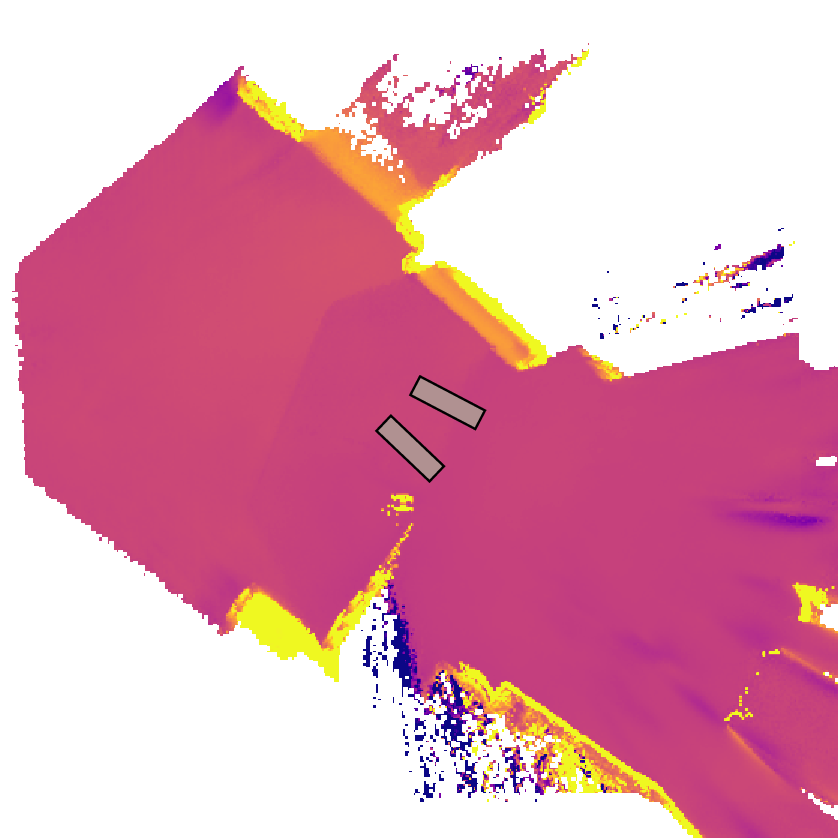}
    \end{tabular}
    \vspace{-2mm}
    \caption*{\small \bf EKF-ICP-PC}
    \vspace{2mm}

    \begin{tabular}{cc}
        \includegraphics[width=0.20\textwidth]{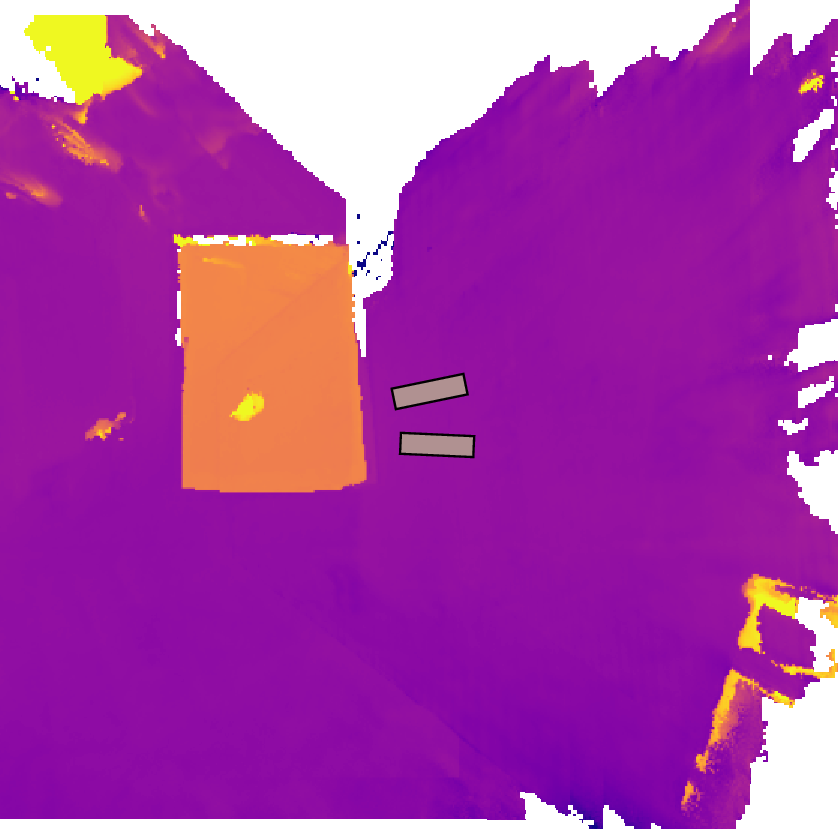} &
        \includegraphics[width=0.20\textwidth]{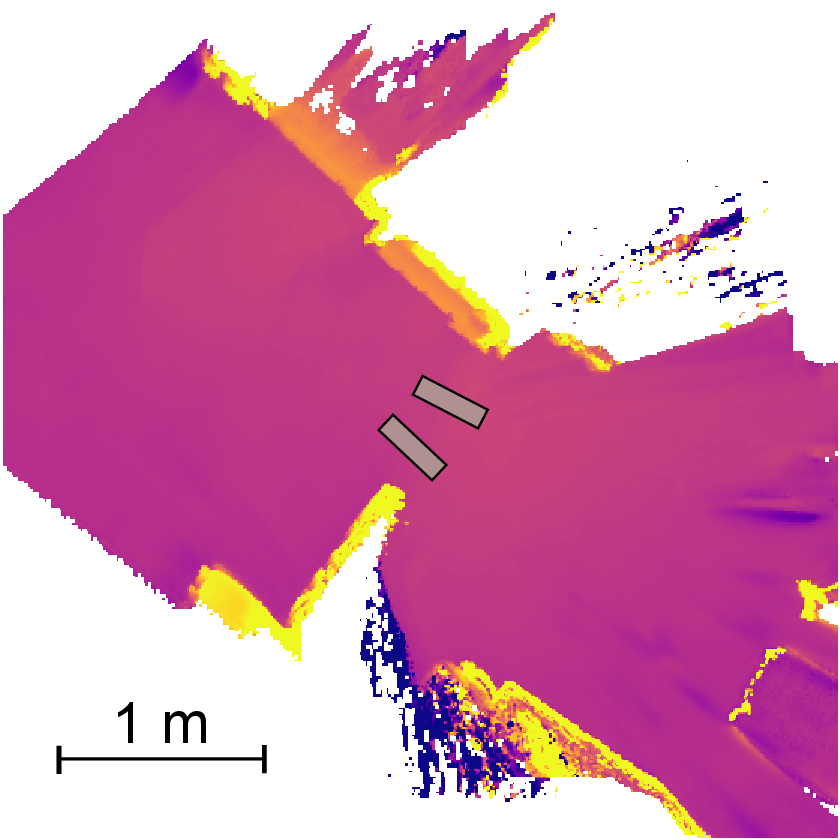}
    \end{tabular}
    \vspace{-2mm}
    \caption*{\small \bf EKF-ICP-EM}
    
    \caption{Elevation maps from both experiments. The three maps from each experiment use the same color scale. The zero elevation level corresponds to the height of the feet, which are represented by the gray rectangles.}
    \vspace{-3ex}
    \label{fig:elevation maps}
\end{figure}

The right column of Fig. \ref{fig:elevation maps} displays the elevation maps as the exoskeleton passes through a narrow doorway. To avoid colliding with the door frames, several small in-place steps were made. With each step, EKF-P accumulates minor errors in the vertical direction, resulting in the floor in front of the exoskeleton being mapped at progressively higher levels. This is evident from the color difference between the floor directly in front of and behind the exoskeleton. In contrast, the ICP-based estimators effectively correct these errors, maintaining consistent elevation levels on both sides of the doorway.

\begin{table}[t]
    \centering
    \small
    \caption{Relative Error (RE) and Absolute Trajectory Error (ATE) metrics for translation (cm) and rotation (º) in experiment 1. RE values represent the median of the relative errors over 4-meter windows.}
    \begin{tabularx}{0.9\columnwidth}{cYYYY}
        & \multicolumn{2}{c}{\textbf{RE}} & \multicolumn{2}{c}{\textbf{ATE}} \\
        \cmidrule(lr){2-3} \cmidrule(l){4-5}
                                       & \textbf{Trans.} & \textbf{Rot.} & \textbf{Trans.} & \textbf{Rot.}    \\
        \midrule
        \textbf{EKF-P}       & 2.08          & 1.40          & 5.27          & 1.58            \\
        \textbf{EKF-ICP-PC}  & 2.08          & 1.38          & 5.29          & 2.80            \\
        \textbf{EKF-ICP-EM*} & 1.52          & 1.13          & 3.93          & 2.03            \\
        \textbf{EKF-ICP-EM}  & \textbf{1.49} & \textbf{1.11} & \textbf{2.93} & \textbf{1.44}   \\
        \bottomrule
    \end{tabularx}
    \vspace{-5mm}
    \label{tab:metrics}
\end{table}

\subsection{Discussion}

The two experiments performed demonstrate that EKF-ICP-EM enhances both the accuracy of the odometry and the smoothness of the elevation maps compared to the purely proprioceptive estimator EKF-P. Although both methods exhibit some vertical drift, a closer analysis reveals that their underlying sources differ. In EKF-P, drift mostly comes from the accumulation of errors in estimating the height at which the feet contact the ground after each step, resulting in maps with elevation discontinuities. In contrast, drift in EKF-ICP-EM primarily occurs when the registration procedure introduces a slight inclination error in the elevation map. As a result, subsequent registrations cause the trajectory estimates to follow a nonexistent slope, while the map remains smooth.

Among the two ICP-based methods, EKF-ICP-EM has a clear advantage in that it eliminates the need to construct and update an additional representation of the environment. Instead, it operates directly on the elevation map, which would already be generated and maintained for the exoskeleton to plan its movements. Moreover, registering new point clouds on the elevation map is significantly faster than using a point cloud map, as both the nearest neighbor search and the computation of normal vectors are more efficient on elevation maps. In our implementation, the elevation map-based ICP variant runs five times faster than the point cloud map-based alternative\footnote{157 ms vs 773 ms in average per frame, measured on laptop with an AMD Ryzen 5 3500U CPU with a single-threaded implementation in Python with C++ bindings.}.

Since a point cloud can capture more detailed information of the environment than an elevation map, one might expect EKF-ICP-PC to consistently outperform EKF-ICP-EM. However, both experiments suggest that the trajectory estimates of EKF-ICP-PC are actually less accurate. This counterintuitive result can be explained by the fact that, given the sensor placement on the exoskeleton, the small region of the ground visible to the depth camera is often so geometrically simple that representing it as a point cloud or an elevation map makes little difference. Moreover, because the elevation map naturally filters out depth noise over time, it maintains a smoother representation of the ground compared to the point cloud map, which retains noisy depth measurements. As a consequence, the normal vectors from the elevation map are in general more reliable than the ones from the point cloud map. Our experiments indicate that, under these visibility conditions, the benefits of the elevation map outweigh its drawbacks, making EKF-ICP-EM the better choice for this application.

\section{CONCLUSION}

In this paper, we presented a new odometry algorithm specifically designed for leg exoskeletons equipped with a limited FOV depth sensor, which combines a proprioceptive state estimator with a novel elevation map-based ICP variant. Our approach produces more accurate trajectory estimates and elevation maps compared to the purely proprioceptive method, and is able to handle the underconstrained scenarios that would normally be challenging for point cloud registration. The experiments conducted indicate that our ICP variant not only improves computational efficiency but also outperforms a point cloud map-based algorithm, despite relying on a simpler representation of the environment.

\addtolength{\textheight}{-13.4cm}  




\section*{ACKNOWLEDGMENT}

The authors would like to thank Jianeng Wang and Matias Mattamala of the Oxford Robotics Institute for creating the point cloud map used as basis for the floor plan in Fig. \ref{fig:trajectories_2}.

\bibliographystyle{IEEEtran}
\bibliography{references.bib}

\end{document}